%% file: main.tex
\documentclass[sigconf]{acmart}

\usepackage{soul}
\usepackage{bookmark}
\usepackage{amsfonts}
\usepackage{enumitem}

\usepackage[ruled,vlined,linesnumbered]{algorithm2e}
\usepackage{mathtools}
\usepackage{subcaption}
\usepackage{csquotes}
\setlist[itemize]{leftmargin=*}
\setlist[enumerate]{leftmargin=*}

\graphicspath{ {./figures/} }

\AtBeginDocument{%
  \providecommand\BibTeX{{%
    \normalfont B\kern-0.5em{\scshape i\kern-0.25em b}\kern-0.8em\TeX}}}

\setcopyright{acmcopyright}
\copyrightyear{2023}
\acmYear{2023}
\acmDOI{}

\acmConference[MLF, KDD '23]{KDD '23}{August 6--10, 2023}{Long Beach CA, USA}
\acmBooktitle{Workshop on Machine Learning in Finance}
\acmPrice{}
\acmISBN{}

\begin{document}

\settopmatter{printfolios=true}
\title{Adversarial training for tabular data with attack propagation}

\author{Tiago Leon Melo}
\email{tleonmelo@hotmail.com}
\authornote{Work developed while employed at Feedzai.}
\affiliation{%
  \country{Feedzai}
}
\author{João Bravo}
\email{joao.bravo@feedzai.com}
\affiliation{%
  \country{Feedzai}
}

\author{Marco O. P. Sampaio}
\email{marco.sampaio@feedzai.com}
\affiliation{%
   \country{Feedzai}
}

\author{Paolo Romano}
\email{romano@inesc-id.pt}
\affiliation{%
   \country{INESC-ID and IST, U. de Lisboa}
}

\author{Hugo Ferreira}
\email{hugo.ferreira@feedzai.com}
\affiliation{%
  \country{Feedzai}
}

\author{Jo\~ao Tiago Ascens\~ao}
\authornotemark[1]
\email{jtascensao@gmail.com}
\affiliation{%
   \country{Feedzai}
}

\author{Pedro Bizarro}
\email{pedro.bizarro@feedzai.com}
\affiliation{%
   \country{Feedzai}
}

\renewcommand{\shortauthors}{Melo et al.}

\begin{abstract}

 Adversarial attacks are a major concern in security-centered applications, where malicious actors continuously try to mislead Machine Learning (ML) models into wrongly classifying fraudulent activity as legitimate, whereas system maintainers try to stop them. Adversarially training ML models that are robust against such attacks can prevent business losses and reduce the work load of system maintainers. In such applications data is often tabular and the space available for attackers to manipulate undergoes complex feature engineering transformations, to provide useful signals for model training, to a space attackers cannot access. Thus, we propose a new form of adversarial training where attacks are propagated between the two spaces in the training loop. We then test this method empirically on a real world dataset in the domain of credit card fraud detection. We show that our method can prevent about 30\% performance drops under moderate attacks and is essential under very aggressive attacks,
 with a trade-off loss in performance under no attacks smaller than 7\%. 

\end{abstract}





\begin{teaserfigure}
\vspace{2mm}
\begin{center}
\includegraphics[width=0.85\linewidth]{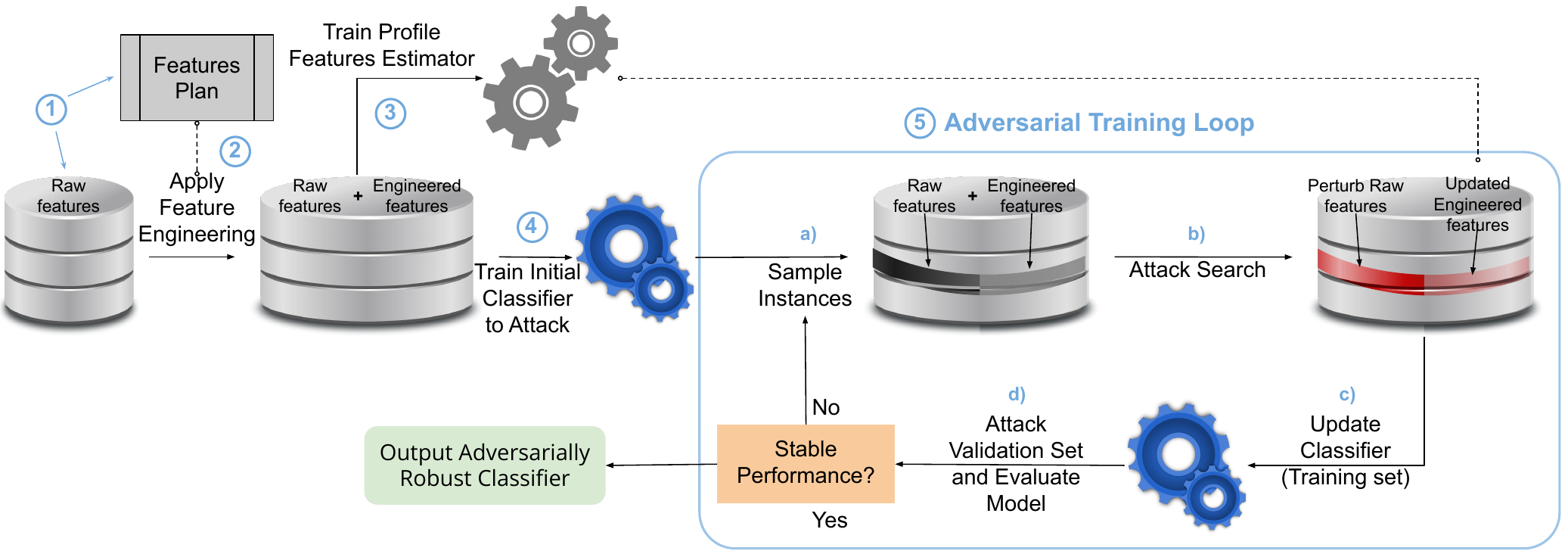}
  \caption{Adversarial Training System for Tabular Data with Attack Propagation (see summary in Sect.~\ref{sec:methods})}
  \label{fig:overall_system_diagram}
\end{center}
\end{teaserfigure}

\maketitle
\input{introduction}
\input{methods}

\input{experiments}

\input{related_work}

\input{conclusions}
\begin{acks}
This research has been supported by projects with references \\ UIDB/50021/2020, POCI-01-0247-FEDER-045915 and \\ C645008882-00000055.PRR.
\end{acks}


\bibliographystyle{ACM-Reference-Format}
\bibliography{references}

\end{document}

%% file: introduction.tex
\section{Introduction}\label{sec:introduction}

Machine Learning (ML) models are vital in many security-sensitive applications \cite{papernot2016distillation}. In the financial services industry they are used to classify, for example, credit card transactions as legitimate or fraudulent~\cite{cartella2021adversarial}. In this naturally adversarial setting, fraudsters continuously adapt their techniques to bypass the system, while the system maintainers try to stop them. Thus, being able to train a stable model that withstands such attacks is valuable to avoid frequent and expensive model retraining operations with fresher data.

In tabular data domains, the raw features space containing some of the features available for the attackers to manipulate directly is often enriched, for model training, into an enlarged space via complex temporal and entity based aggregations~\cite{rnns_feedzai_10.1145/3394486.3403361}. This poses the additional challenge of incorporating the propagation of perturbations between two spaces in the adversarial training loop. Furthermore, designing appropriate attacks and attack search methods is essential to obtain models that are robust. Thus we propose an extension of adversarial training with all these ingredients
and perform an empirical study on a credit card fraud detection dataset. Methods that tackle a similar setting do not address the propagation of perturbations and focus only on attacks~\cite{cartella2021adversarial}. 

Our main contributions are as follows:
\begin{itemize}
    \item We parameterize adversarial perturbations and develop methods to search for strong attacks on tabular data (Sect.~\ref{subsec:adversarial-perturbations},~\ref{sub:pert_norm} and~\ref{subsec:perturbation-search-strategies}). 
    \item We formulate different approximation methods to efficiently update complex temporal aggregations as a result of adversarial perturbations of raw features (Sect.~\ref{subsubsec:perturbed-features-update-methods}).
    \item We introduce an adversarial training framework that increase the robustness of a classifier against a broad range of attacks (Sect.~\ref{subsec:adversarial-training}). Particularly, in our empirical study (Sect.~\ref{sec:experiments}) we develop a model that is well protected against multiple attack strategies generalizing well under new attacks in test. This is achieved at a cost in performance on test data without attacks between $4\%$ and $7\%$, (the adversarial robustness trade-off~\cite{tsipras2019robustness,zhang2019theoretically}), while avoiding disastrous losses if the developed attacks arise (from $30\%$ loss to random guessing performance, depending on the perturbations).
\end{itemize}
The paper is structured starting with a description of the methods (Sect.~\ref{sec:methods}) followed by the empirical study (Sect.~\ref{sec:experiments}), and a brief summary of related work (Sect.~\ref{sec:related-work}) before the conclusions (Sect.~\ref{sec:conclusions}).

%% file: methods.tex
\section{Methods}
\label{sec:methods}

Figure~\ref{fig:overall_system_diagram} shows a high level view of our adversarial training method. We summarize the main parts leaving details for later sections:

\begin{enumerate}
    \item \textbf{Inputs}:  
    \begin{itemize}
        \item \textit{Raw features dataset (Section~\ref{subsubsec-input-data})}: We assume we have a tabular dataset with $N$ rows as input. For each row, the vector of raw feature values to be perturbed is $x \in \mathcal{X}$ and the classification target values is $y\in \mathcal{Y}$.
        The union of all rows composing the dataset is denoted by matrices $X \in \mathcal{X}^N$ and $Y \in \mathcal{Y}^N$. 

        \item \textit{Features plan (Section~\ref{subsubsec-input-data})}: We assume the set of raw features is enriched via a transformation $f: \mathcal{X}^N \to \mathcal{X}'$ of the input rows $X$ to produce a new vector of features $x' = f(X)$ for each row. The matrix of new feature values for all rows is denoted by $X'$. 
        A row-by-row mapping operation is a special case where, instead, $x\to x' = f(x): \mathcal{X} \to \mathcal{X}'$. 
        Feature engineering is important 
        in tabular data domains
        and it often involves non-bijective mappings from the raw dataset, which forces us to adapt the adversarial training loop. Aggregations that group several instances, such as time window aggregations grouped by an entity (e.g., user), which we denote as profiles, and feature extraction from text fields, are common examples.
    \end{itemize}
    \item \textbf{Apply feature engineering:} The first step is then to enrich the original raw features dataset by applying the features plan.
    \item \textbf{Train Profile Features Estimator} (Section~\ref{subsubsec:perturbed-features-update-methods}): 
    Profile feature re-computations can be expensive inside the adversarial training loop. For a faster adversarial training we train auxiliary models to estimate changes in profiles under adversarial attacks, using only the unperturbed event.
    \item \textbf{Train Initial Classifier (to be made robust):} Before adversarial training, we train a model on the unperturbed dataset. 
    \item \textbf{Adversarial Training Loop} (Section~\ref{subsec:adversarial-training}):  This consists of several rounds of i) searching for attacks on a portion of the training set instances and ii) updating the model by including those perturbed instances in the training. The main goal is to obtain a robust model that, after a number of rounds, performs better than the initial model trained without attacks, when both are subject to the same attack search strategy. Our steps of adversarial training are as follows: 
    \begin{enumerate}
        \item \textit{Sample Instances:} We focus on a binary classification use-case where only the positive instances are sampled to be attacked. The goal is to lead the model into wrongly classifying positives as negatives.
        \item \textit{Attack Search} (Section~\ref{subsec:adversarial-perturbations}): Next we search for the best attack for each of the selected instances. Our attacks are targeted at reducing the positive class score by perturbing a set of raw fields. Then the perturbations are propagated to the engineered features. In particular, the perturbations of the profile features are applied using the pre-trained \textit{Profile Features Estimator}. 
        \item \textit{Update Classifier:} The attacked dataset is then used to tune the model by performing several gradient steps (we use a gradient boosted decision trees algorithm).
        \item \textit{Attack Validation Set 
 and Evaluate Model:} Finally, the validation set is attacked and a target performance is evaluated on it to verify if the model performance stabilized or if adversarial training continues.
    \end{enumerate}
\end{enumerate}
In the next sections we provide further details of the method.

\subsection{Input Data Transformations}
\label{subsubsec-input-data}

We assume the input data contains numerical features, text fields and categorical features.
Complex entities may be present combining various features. For example, a card entity combines a categorical feature to identify the credit card network with a card number and expiry date. 
So, an attack where a card is switched likely changes many attributes at once. 
Next we describe transformations to produce features in $\mathcal{X}'$.

\subsubsection{Row-wise map operations:} 
Simple examples in transaction fraud detection are: i) compute a ratio between two features, ii) extract a country category from a phone number, iii) extract geolocation from a zip code, etc...

\subsubsection{Aggregations of groups of rows:}
These features encode more complex information.
Examples are temporal or spatial windowed aggregations that group together instances within a time frame or in a given spatial region 
(a count of transactions by a card in the last 24 hours is such a profile feature aggregation). In this example, perturbing the timestamp of the targeted row could potentially influence which rows are used in the aggregation, 
thus making adversarial training non-trivial. On the other hand, if perturbations are applied on the co-domain $\mathcal{X'}$ directly, i.e., on the engineered features, we can obtain values that do not correspond to any configuration of perturbed rows in the domain $\mathcal{X}^N$. Or, even if they do, finding the exact domain values might be very complex, involving finding the exact grouping configuration originating the perturbed feature value.
More specifically, consider the example of the count of transactions by a card in the last 24 hours. If the count is perturbed directly, e.g., from 5 to 12, we have to find how to adjust the timestamps of several other rows so that they fall into the last 24 hours to increase the count (with the complex side effect of perturbing several other groups containing the adjusted rows). This example shows that devising an efficient way to propagate perturbations from $\mathcal{X}$ to $\mathcal{X}'$ is key for adversarial training (later discussed in Section~\ref{subsubsec:perturbed-features-update-methods}).

\subsubsection{Higher order transformations:}
Finally, it is common to apply secondary transformations to the generated features. This is useful to provide direct signals to the model.
For example, after computing a profile feature for the average and standard deviation of the amount spent by a card in the last week, compute a z-score for the amount of the current transaction to signal how much of an outlier the observed amount is. 


\subsection{Adversarial Perturbations}
\label{subsec:adversarial-perturbations}
We discuss how to define adversarial perturbations on $\mathcal{X}$, propagate them to $\mathcal{X}'$, measure their magnitude and search for them.

\subsubsection{Types of perturbations}
\begin{itemize}
    \item \textit{Categorical perturbations:} Categorical features are perturbed by replacing the category value by an existing value.
    \item \textit{Text perturbations:} The string in a text feature can be changed in part (e.g., replacing, removing or adding characters) or fully. If used as a categorical it amounts to replacing the category value. If used for feature extraction then derived features can be affected.
    \item \textit{Numerical perturbations:} These can be defined by a shift, scaling or any other numerical mapping. 
    \item \textit{Grouped perturbations:} In some cases groups of features may be perturbed together in a correlated way, either categorical, text or numerical groups or mixtures of any of them. For example, a perturbation of a latitude coordinate may imply a perturbation of the longitude within the constraint of not falling into the ocean.
\end{itemize}

\subsubsection{Perturbation Propagation Methods}
\label{subsubsec:perturbed-features-update-methods}

In this section, we start by discussing how perturbations propagate from $\mathcal{X}$ to $\mathcal{X}'$, in the credit card fraud detection use case, to present our perturbations pipeline. Then we discuss practical methods to update such derived features either exactly or approximately.

\subsubsection*{Examples of effects of perturbations on engineered features}
\begin{figure*}
\centering
\includegraphics[width=0.8\textwidth]{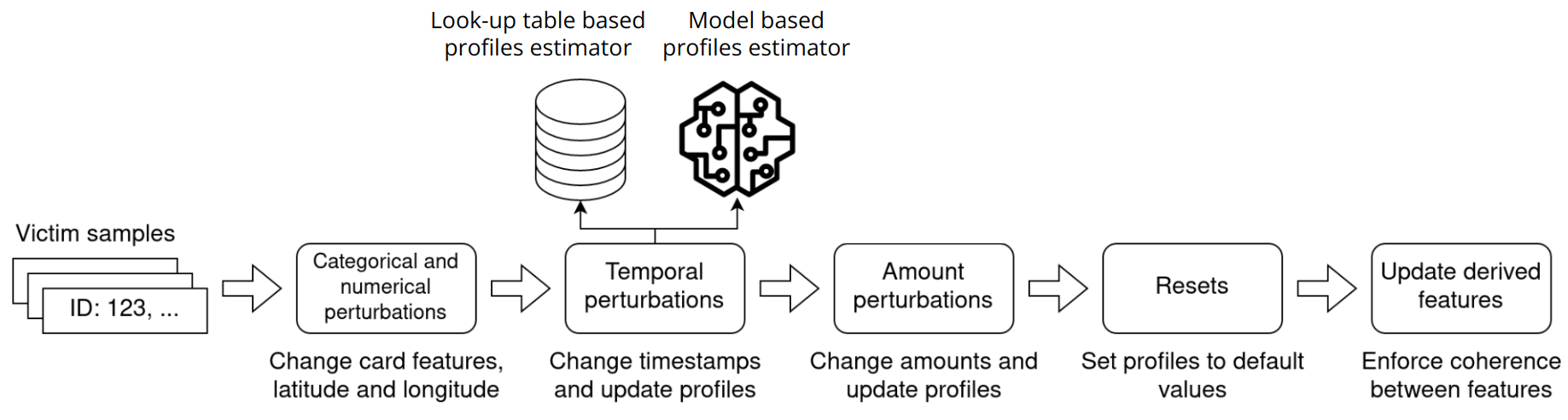}
\caption{Diagram for the full perturbation pipeline}
\label{fig:pipeline}
\end{figure*}

In credit card fraud detection common attacks by fraudsters are as follows:
\begin{itemize}
    \item \textit{Amount perturbations} -  
    A fraudster can alter the numerical amount of a transaction, thereby affecting profile features.
    \item \textit{Temporal perturbations} - This impacts profile features. Shifting the timestamp of an instance alters which instances are in the time window, thereby affecting the aggregation outcome.
    \item \textit{Card resets or switches} - 
    In fraud detection
    this corresponds to a fraudster switching a card that was identified as stolen. The engineered features involving the card identifier as a grouping criterion are reset to a base value (e.g., the count of transactions of the card in a time window goes back to 1). If all the card's features are changed (e.g., the card issuing network also changes) we call it a card switch.
    \item \textit{Geolocation perturbations} - This is an example of a group of numerical features (latitude and longitude) that are typically perturbed simultaneously. In some scenarios, fraudsters could spoof their IP, which is used to identify their location. 
\end{itemize}
In our method we apply such perturbations to a transaction in 4 steps: i) change categorical and numerical features directly (either individually or in groups) in an exact manner, ii) shift timestamp and update the corresponding profiles, iii) change the amount and update profiles 
and iv) reset profiles by changing the values of grouping entities (e.g., card). A representation of this pipeline applied to a perturbed (victim) instance is displayed in Figure~\ref{fig:pipeline}.

\subsubsection*{Exact updates}
We now provide examples of exact updates of the engineered features in $\mathcal{X}'$ following the pipeline of Figure~\ref{fig:pipeline}:
\begin{itemize}
    \item \textit{Categorical features updates:} 
    To perturb such a feature one simply changes its value (e.g., switch the card issuer from Visa to Mastercard).
    When perturbing categorical features, it is important to note that they are frequently interdependent
    For instance, the distribution of Card Verification Value (CVV) flags may differ across various card issuers. Therefore, some categorical feature groups have to be perturbed together. Thus, for a card switch, we replace the values of its features by those of another card, selected randomly with uniform probability (i.e., reproducing the original dataset's distribution).
    Similar arguments hold for network features. After categorical features are perturbed, other dependent features may also be updated either exactly or approximately (see other examples provided in this and the next section).
    \item \textit{Latitude and Longitude:}
    Values are selected for both coordinates simultaneously using a uniform distribution from regions with sufficient density in the original dataset.
    \item \textit{Aggregation function value updates}: Even if a group of instances used in an aggregation does not change (e.g., as a result of a timestamp perturbation), other perturbed features may change the output of the aggregation. For sum profiles and other associative operations this can be updated exactly by computing differences. For example, if the sum amount in the last hour for a victim card is 1200 when the amount for the current transaction is 100, the new sum amount will be 1150 when the transaction amount is perturbed to 50.  
    \item \textit{Higher order transformation updates:} Other secondary feature updates are, in many cases, straightforward, since most consist of ratios between profiles or derived features via functions of a single row. 
\end{itemize}

\subsubsection*{Approximate updates}

We now discuss methods to approximate temporal aggregations (i.e., profiles):

\begin{itemize}
    \item \textit{Data-driven}: For time windows with a large number of instances in each group (high volume profiles), we estimate perturbations of a single instance using a look-up table that stores the mean profile value over time for each profile. The assumption is that 
    the main source of change is not due to the instance shifting in time but, instead, is due to the group of instances changing.  An example of a high volume profile is  the number of transactions for a given merchant (e.g., an electronics retail website) in the last month.
    For these profiles we compute a histogram of values by binning over time (e.g., calculate the mean for each profile using bins of one hour - see diagram in Figure~\ref{fig:dd-diagram}). 
    \begin{figure}[h]
    \centering
    \includegraphics[width=\linewidth]{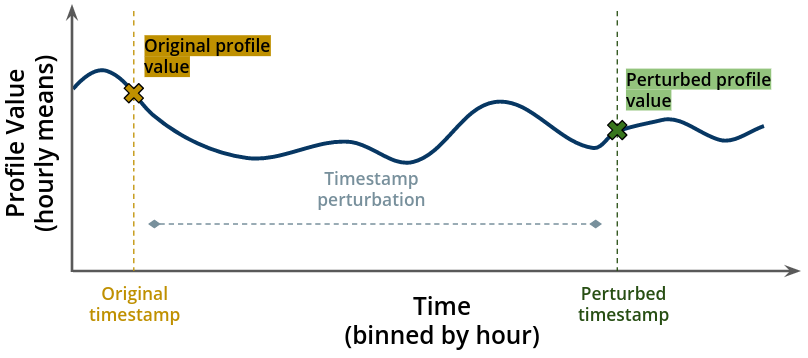}
    \caption{Profile estimation for perturbations or large groups.}
    \label{fig:dd-diagram}
    \end{figure}
    Then, for the perturbed value of a profile feature, we use the mean value found in the look-up table for this feature at the new temporal bin (due to the time shift). 
    This approach assumes that within the time-frame of 1 bin, the profile value does not change significantly around the mean when a single event is inserted or removed. 
    \item \textit{Model-based}: For profiles with a small number of instances in each group (low volume) we estimate their changes over time by fitting a multi-output regression model. To train each model, we build a dataset containing instances perturbed several times, corresponding to profile changes for various perturbation time intervals (positive or negative delays). The features of one perturbed example, for the regression task, consist of the  raw and engineered features before the perturbation (including profiles) and the amount of delay that was injected in the timestamp. The regression targets are the resulting profile values computed exactly. We consider delays between one minute and one week for consistency with the typical time windows used for profiling. We choose to draw delays $\delta t$ from a logarithmic (scale invariant) distribution with constant probability for $\log |\delta t|$, and with equal probability for either sign.

\end{itemize}  

\subsection{Perturbation Norms}
\label{sub:pert_norm}
To be able to quantify the magnitude of a perturbation, we define a norm on $\mathcal{X}$, which is specifically tailored to the use case we want to investigate. In credit card fraud detection, fraudsters are only able to manipulate the raw features directly so a norm on this space reflects the real costs driving the fraudster's decisions when attacking the model. Furthermore, a norm on the raw features is more practical, since it can be computed before the (often complex) updates of the engineered features. 
It is defined by assigning a custom set of costs for each attacked feature
and then summing them.  To gain insights on the relative importance of each type of perturbation, we interviewed domain experts whose knowledge helped defining the following parametrization of perturbations:
%
%
\begin{itemize}
\label{list:vectors}
    \item \textit{Amount perturbations cost}: We allow to vary between 2\% to five times larger than the original value through a scaling factor. We use a custom cost function of the scaling factor $s$ that grows as:
    \begin{equation}
        \sim c_{\textrm{amount}} \begin{cases}
        \log(s) &, s > 1 \\
        \log(1/s) &, s \leq 1
        \end{cases}
    \end{equation}
    where $c_{\textrm{amount}}$ is a constant. Note that, from a fraudster's perspective, both reducing or increasing the amount incurs a penalty related to trial and error (increasing the risk of blocking several cards while trying). Increasing the amount offsets some of the penalty with extra profits from using a larger amount. In contrast, reducing the amount always reduces the profit a fraudster can extract from a stolen card, which is why a rapidly growing penalty is used when $s$ decreases.
    %
    \item \textit{Temporal perturbations} - We allow time shifts $|\delta t|$ (in milliseconds) smaller than 1 week with a cost $\sim c_{\mathrm{temporal}} \log(|\delta t| + 1)$.
    \item \textit{Card resets and switches} - It is frequent for a fraudster to have access to various stolen cards (e.g., stolen online banking details that allow them to generate virtual cards). We consider two types of perturbations: i) card resets, which consist of preserving card features (e.g., name, issuing network, etc...) while only changing the card id, ii) card switches, which consist of using a completely new card with all features changed (assumed to cost more).
    \item \textit{Geolocation and Network perturbations} - Fraudsters can easily manipulate these (e.g., changing IP through a Virtual Private Network). Thus, we assign small constant costs to each of these.
\end{itemize}
A summary of the costs used in the experiments is provided in Table~\ref{tab:costs}.
\begin{table*}
\centering
\begin{tabular}{|r|c|c|c|c|c|c|}
\hline
\textbf{Perturbation} & Network & Geolocation & Temporal & Amount & Card reset & Card switch (=Card Reset + Additional Cost)    \\ \hline \hline
\textbf{Cost}         & 3       & 4           & 18 ($\max$)       & 26 ($\max$)     & 33         & 49 (= 33 + 16) \\ \hline
\end{tabular}
\caption{Cost of each perturbation used in the experiments.}
\label{tab:costs}
\end{table*}
We normalize the values so that the norm ranges from 0 to 100 (when all maximal perturbations are applied). 
It is important to note that the choice of norm is use-case specific. The values we choose for the experiments, though well motivated, are illustrative.

\subsection{Perturbation Search Strategies}
\label{subsec:perturbation-search-strategies}
We now describe methods to generate attacks on $\mathcal{X}$. The goal is to generate attacks with the highest possible success rate under the lowest possible search budget.  The main attack generation strategies in the literature are: \textit{black box}, \textit{white box} and \textit{white box proxies}, respectively corresponding to attacks with access to: i) the system decision, ii) the full model, iii) to a model that imitates the true model. Another intermediate possibility, which we explore, is to attack a black box model in a partial observability setting, i.e., with full access to the model outputs (e.g., model scores and not only the final decision). 

\subsubsection*{Random Search}
This is a black box approach that serves as a baseline for the other search methods.
For each sample to be perturbed, many random perturbations are generated independently by sampling a Bernoulli variable for each feature, to decide if the feature is perturbed.  The Bernoulli trial probability is kept low so that it is not very likely to perturb an instance using all raw perturbations, which may skew the distribution of norms to large values. For each perturbed feature, a new value is generated from a feature specific distribution (as discussed in Sect.~\ref{subsubsec:perturbed-features-update-methods}).
A similar strategy is used for every independently perturbed feature, whereas for features groups (e.g., card features) a single Bernoulli trial is used to perturb the group.
%

To evaluate an adversarial strategy, such as random search, we can consider its success rate in addition to the size of the perturbations it generates. The success rate is computed as:
\begin{equation}
    \text{Sucess Rate} = \frac{\# \; \text{successful attacks}}{\# \; \text{generated attacks}}\; .
    \label{eq:sr}
\end{equation}
For random search, an adversarial attack is successful if any of the generated attacks on an instance switches the model decision. 

\subsubsection*{Stochastic Coordinate Descent}
Given the high prevalence of discrete variables in the search space for the use case at hand, we employ standard algorithms from the discrete optimization literature, namely Stochastic Coordinate Descent (SCD)~\cite{scd}. SCD is initialized with a clean input instance followed by several iterations sweeping over various features. A sweep consists of repeatedly picking a different random feature to attack until all features have been selected. For each selected feature we explore its values, and update the perturbation value if it is better than the last perturbation (e.g., in fraud detection if it results in a lower model score, corresponding to a lower fraud risk). In this method convergence occurs whenever a full sweep of all features yields no improvements. In particular, whenever a feature is selected, a grid of possible values to explore is generated within the norm constraint on the perturbation. For categorical features all values are explored whereas for numerical features a discrete grid of values is generated. 

Variations of this method can be formulated depending on the criterion used to select the best perturbation from the grid, 
namely:
\begin{itemize}
    \item \textit{Greedy approach}: When the goal is to trick the model into classifying positives as negatives, we accept perturbations yielding a smaller model score.

    \item \textit{Cost-efficient approach}:
    In each attempt to perturb a feature, we select the perturbation with the best norm change to score change ratio. Thus, this approach tries to avoid getting stuck in local minima by optimizing for cost-efficiency. 
\end{itemize}

\subsubsection*{Greedy Search}
Another approach, which tries to circumvent the local minima problem, consists of generating and exploring a grid for every feature on each iteration and pick the best feature to perturb.
The success rate of this method 
is expected to be higher at the expense of more model score evaluations and, thus, slower convergence.

\subsection{Adversarial Training}
\label{subsec:adversarial-training}
In this section we describe specific choices of our implementation of adversarial training described at the beginning of Section~\ref{sec:methods}.

We begin by generating attacks against a baseline model. These attacks take a percentage of the positives in the training set and all positives in the validation set. The attack strategy has its own set of parameters. The norm constraint is the most important one, as it determines the maximum cost of the attack and significantly affects the attack's success rate.

After replacing the generated samples in the training and validation sets we update the model by running several iterations of gradient boosting with early stopping.

Finally, to validate the updated model we generate new attacks on the validation set and compute three metrics. We evaluate the model's performance against a dataset without any adversarial attacks with the \textit{Clean pAUC at a 1\% FPR}, i.e., the area under the ROC curve up to $1\%$ False Positive Rate\footnote{In fraud detection, the operating region is usually at a low FPR to keep the number of alerts under control.}. 
This gives us an unbiased estimate of its performance 
on clean test set. 
Similarly, we compute the \textit{Adversarial pAUC at 1\% FPR} 
on the attacked version of the validation set, which can only be as high as the clean pAUC, since the attack strategy never generates perturbations that are easier to classify. Lastly, we also assess the attack's \textit{Success Rate}.

%% file: experiments.tex
\section{Experiments}\label{sec:experiments}
In this section we present our empirical study of the adversarial training method. In Section~\ref{subsec:data-preparation} we provide an overview of the datasets and its processing. 
Next, we describe the training of the input models, namely the baseline classifier (Section~\ref{subsec:baseline-classifier}) and the profile estimation models (Section~\ref{subsec:profile-estimation-models}). Finally, we discuss the results of the experiments conducted using adversarial methods, specifically addressing the benchmarking of attacks (Section~\ref{subsec:adversarial-attacks-benchmarking}) and the adversarial training experiments (Section~\ref{subsec:adversarial-training-experiments}).

\subsection{Data Preparation}
\label{subsec:data-preparation}

We use a proprietary dataset of a payment processing network for the task of credit card fraud detection. 
The dataset was sampled by card entity (the main grouping entity for profiles) including 100\% of the cards with fraudulent transaction and 5\% of the cards with exclusively legitimate transactions. After sampling, the resulting dataset contains $\sim$180 million transactions and a fraud rate of 0.29\%.

About 250 of the engineered features were designed by domain experts for a production system, including profiles by card and merchant, row mappings and higher order transformations (see Section~\ref{subsubsec-input-data}). One raw feature flags each transaction as Card Present (CP) or Card Not Present (CNP). We focus on CNP transactions since online transactions are more prone to fraud. After computing the engineered features, we only kept the CNP transactions ($\sim$34 million) with a fraud rate of 1.2\%. The data was split, with the first 10 weeks for training, followed by 4 weeks for validation and 6 weeks for testing.

\subsection{Baseline Classifier}
\label{subsec:baseline-classifier}
Tree-based models, such as gradient boosted decision trees, have been shown to perform better on tabular data in comparison to deep learning approaches \cite{https://doi.org/10.48550/arxiv.2207.08815}.
A state of the art implementation is given by the LightGBM library~\cite{ke2017lightgbm}, which we use to hyperparameter tune and train the baseline classifier on the clean train and validation sets without adversarial attacks. To assess its performance we compute the normalized partial AUC up to an FPR of 1\% and the recall at a fixed FPR of 1\% obtaining, respectively, 0.42 and 0.58.

\subsection{Profile Estimation Models}
\label{subsec:profile-estimation-models}

For the profile estimation models, we employ LightGBMRegressor models using the train and validation sets and evaluate them on the test set. For simplicity, hyperparameter tuning is done for a single profile regression model and then the hyperparameters are fixed to train the remaining models.

\begin{figure}
    \centering
    \includegraphics[width=0.45\textwidth]{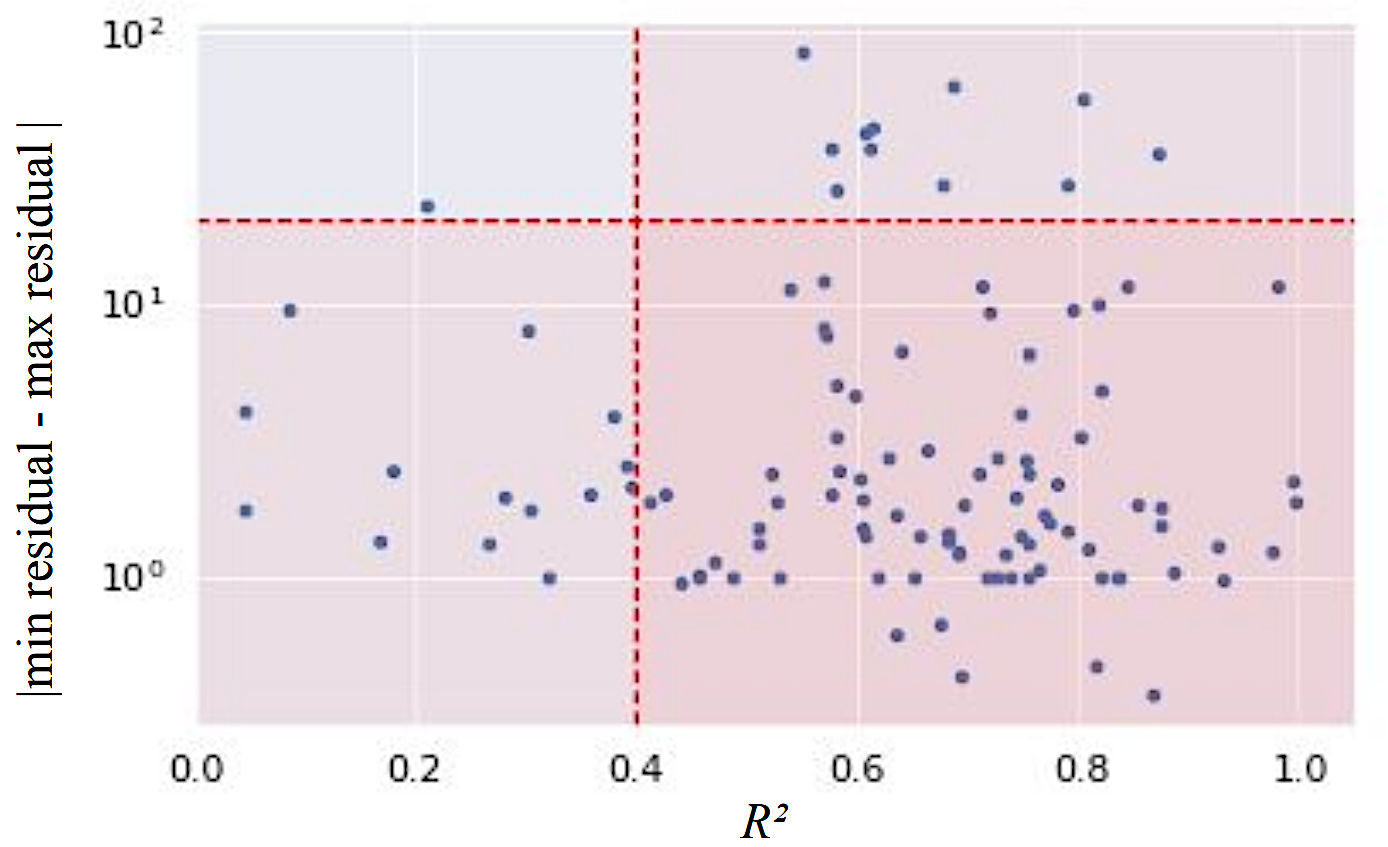}
    \caption{Region of interest for profile predictions. We discard profiles with large normalized residuals and small $R^2$.}
\label{fig:residual_x_r2}
\end{figure}

We evaluated the regression models using the $R^2$ metric. 
We compared the results with a baseline that always keeps the profiles unchanged under the perturbation and obtained large values for $R^2$ but also low ones. 
To identify under-performers, in Figure~\ref{fig:residual_x_r2} we show the absolute difference between the maximum and the minimum residuals for each profile estimator (normalized by the profiles's standard deviation). We highlight the region of interest in darker red (larger $R^2$ and smaller residuals). 
We checked that by removing  the poorly estimated profile features (outside this region) from the training of the baseline classifier, there is no significant drop in the baseline classifier performance. Thus we discard these under-performing profiles in the remainder.

\subsection{Adversarial attacks benchmarking}
\label{subsec:adversarial-attacks-benchmarking}
In this section we compare the various attack strategies by analyzing their success rate \textit{versus} norm constraint trade-offs.
\begin{figure*}
\centering
\includegraphics[width=0.45\textwidth]{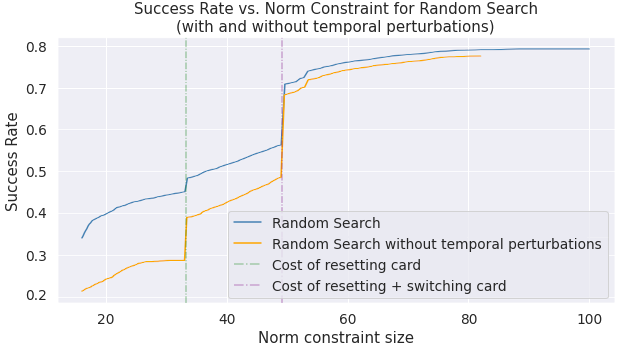}\hspace{0.08\linewidth}\includegraphics[width=0.45\textwidth]{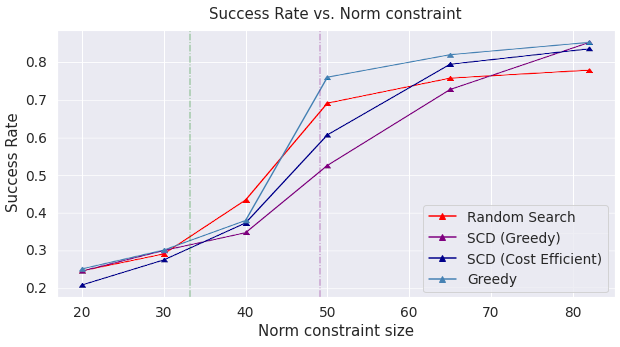}
\caption{Attack Search Strategies Benchmarking: Comparisons of success rate vs norm constraint for  random search with and without temporal perturbations (left) and for all attack strategies without temporal perturbations (right).}
\label{fig:success_rate_plots}
\end{figure*}

\subsubsection{Random search baseline}
In the left plot of Figure~\ref{fig:success_rate_plots} we show the success rate versus the norm constraint for random search. In this benchmark, we ran 500 iterations of Random Search for each case. We display two curves to show the difference in success rates with and without temporal perturbations.
As expected, loosening the norm constraint produces higher success rates, i.e., more successful adversarial attacks. This figure also reveals two very effective discrete perturbations: resetting a card and switching a card (which also demands resetting it), indicated in the figure by the vertical dashed dotted lines. 
Overall, the designed perturbation space can be well explored with a simple random search, finding attacks from as little as 20\% to 80\% success rate.  
The strategy reaches its peak well before it hits the maximum norm constraint (100), suggesting that smarter attack strategies may have space to find successful attacks at large norm constraints.

Before discussing other strategies, we note that the execution times observed for runs with temporal perturbations is four times larger than runs without them. Thus, we opt to ignore them in the remaining attack benchmarks (i.e., the norm will be capped at the corresponding maximum value of 82). In the adversarial training experiments we will, however, re-introduce them.

\subsubsection{Comparison of Black Box Attacks Under Partial Observability}
Regarding the results obtained using the remaining strategies,
%
we show their performance in the right plot of Figure~\ref{fig:success_rate_plots}. The cost-efficient approach produces more efficient attacks than its greedy counterpart for most of the norm constraints. However, both still do not beat the random search baseline except for the largest norm constraint.
The greedy approach, on the other hand, is the best since it beats the benchmark for nearly all norm constraints. 

\subsection{Adversarial training experiments}
\label{subsec:adversarial-training-experiments}
In this section we finally present the results for the full adversarial training experiments. For simplicity, we choose to fix the hyperparameters of the LightGBM algorithm to those found for the baseline classifier. 
Otherwise, we vary parameters related to the adversarial components, namely: the norm constraint used to generate attacks, the fraction of positives to attack in the training set and the frequency of attacks during the model training.




\subsubsection{Parameter Tuning}
\begin{figure*}
    \centering
    \includegraphics[width=0.45\linewidth]{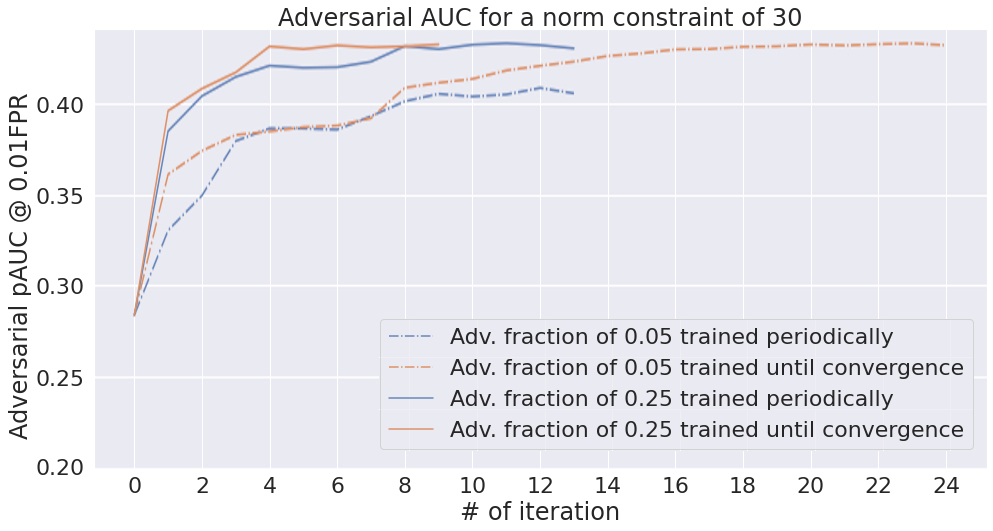}\hspace{0.08\linewidth}\includegraphics[width=0.455\linewidth]{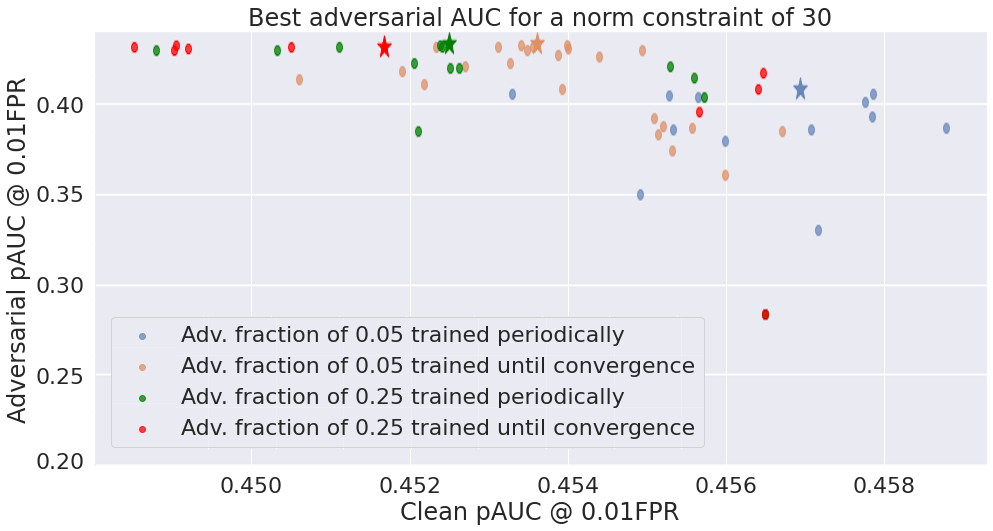}
    \caption{Adversarial Training Evaluations on Validation:  Adversarial pAUC versus the training iteration (left) and versus clean performance (right) for a norm constraint of 30. For each experiment we mark with a star the best pAUC iteration (right).}
\label{fig:adv_auc_val}
\end{figure*}

We start by analyzing the results obtained in validation in four adversarial training configurations and norm constraints of 30 and 65 (respectively not allowing and allowing card perturbations - see Table~\ref{tab:costs}). 
One of the key metrics to indicate if the models are getting more robust is the adversarial pAUC, which we analyze, in Figure~\ref{fig:adv_auc_val}. We display two options for attacking the training set and updating the model: i) attack after a fixed number of gradient boosting model update rounds (i.e., periodically) and ii) attack after full convergence of the gradient boosting model update (evaluated by the pAUC on the validation set).
The results show that larger adversarial fractions in the training set yield faster improvements in adversarial performance in early iterations, though both converge to similar values. Similar conclusions hold for a norm constraint of 65, which we do not display for simplicity. 
%

It is also important to evaluate the performance of the adversarially trained model on clean data. In the right panel of Figure~\ref{fig:adv_auc_val} we visualize the trade-off between clean (horizontal axis) and adversarial (vertical axis) performance at each iteration, for each experiment (the last iteration of each configuration is indicated with a star symbol). 
We observe that clean performance is lower when including a larger adversarial fraction in the training set, as expected. If we consider the best configuration the one with the highest adversarial pAUC, we observe that we only incur a small clean pAUC drop (of about 0.005 pAUC points, i.e., $1.0\%$). Thus, we select the configuration with an adversarial fraction of 0.05 and train until convergence for the remaining experiments. 

\subsubsection{Test Set Results}

In this section, we finally train models with various norm constraints and evaluate them on the test set.
Robustness should improve when stronger attacks are used to train the model, since a model robust against a strong attack is expected to also be robust against a weak one. We test this hypothesis by evaluating the pAUC on an attacked test set for each model.
\begin{figure*}
    \centering
    \includegraphics[width=0.45\linewidth]{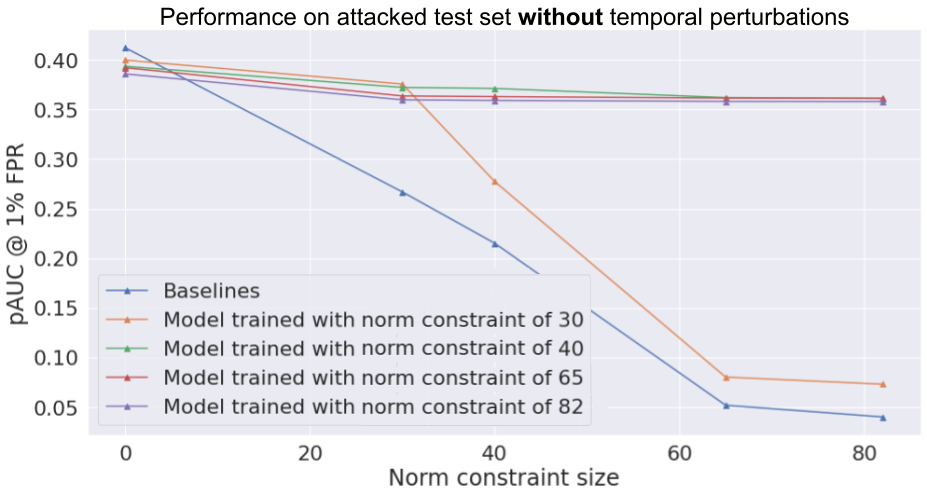}\hspace{0.08\linewidth}\includegraphics[width=0.45\linewidth]{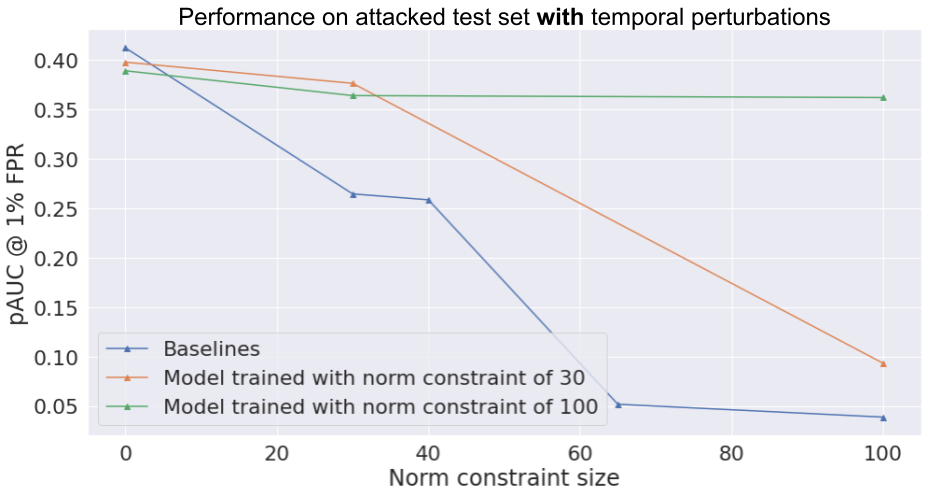}
    \caption{Test Set Evaluations: Adversarial pAUC scores for different norm constraints without (left) and with (right) temporal perturbations included in both training and testing.}
    \label{fig:adv_pauc_norm_a}
\end{figure*}
In the left plot of Figure \ref{fig:adv_pauc_norm_a} we show the adversarial pAUC for models trained with various norm constraints, evaluated on the test set attacked with various norm constraints without temporal perturbations (x-axis).  
We observe that the performance of the baseline model drops fast with increasingly larger attacks, which shows that it is not robust under such attacks.
As for the adversarially trained models, they are significantly more robust. We also see that training with a larger norm constraint does not necessarily produce a more robust model. For example, a model trained with a norm constraint of 65 achieves robustness similar to one with 82. This happens, in part, because both constraints allow the same types of perturbations (see Table~\ref{tab:costs}). 
Also, observe that a model trained with amount and categorical perturbations (norm constraint of 30) is more robust at small norm constraints than models trained with a less constrained norm. This is expected, since a model trained with a more constrained perturbation space should be more optimized to attacks on that space. 


%
In the right plot of Figure~\ref{fig:adv_pauc_norm_a} we show results for experiments with temporal perturbations. Since they are computationally demanding, we only explore two norm constraints, i) only allowing small perturbations up to 30 (i.e., without card resets/switches); and allowing all perturbations (100). 
The overall conclusions are qualitatively similar: as we include larger perturbations we get a model that is more robust to a broader spectrum of perturbations.

Finally, focusing on the points with a norm constraint of zero in the plots of Fig.~\ref{fig:adv_pauc_norm_a} we observe that, 
as expected, we sacrifice some performance on clean data, since the blue point for the baseline evaluated on clean data, is above the adversarially robust models evaluated also on clean data. The observed drop ranges between $4\%$ and $7\%$ (the highest for the models trained with the most aggressive attacks injected).

%% file: related_work.tex
\section{Related Work}
\label{sec:related-work}
\if false
\textbf{Target structure:} focus on the formulation that matters for the methods developed in the thesis
\begin{itemize}
    \item Brief overview of the history of the adversarial attacks and training literature. This may also include mentions to various formulations in a very summarized way.
    \item Maybe focused subsection detailing the formulation that matters for the paper (i.e., attacks that aim at switching the prediction of the model, and adversarial training). Other formulations are mentioned briefly before. (NOTE: we may also want to move details on the formulation to the methods section and keep this ultra short!?).
    \item Subsection on the literature on attacking strategies that are particularly relevant for this work, including the definition of norms (or maybe a different subsection for this point).
    \item Subsection on the literature on adversarial training strategies that are relevant to our method. Here we can also mention other strategies and why they are not suitable for our problem while the one we use is.
\end{itemize}
\fi

Adversarial robustness has recently gained a lot of interest with many studies addressing new strategies to design attacks \cite{goodfellow2015explaining, carlini2017evaluating, cheng2018queryefficient, cartella2021adversarial, hsja, ballet2019imperceptible, permuteattack, https://doi.org/10.48550/arxiv.2203.06414}. 
%
%
The problem is typically stated as follows. Consider the loss function $J(\theta,x,y)$, which corresponds to the cost of classifying an example $x$ of label $y$ with a model parameterized by $\theta$. 
The adversary's goal will often be to attack the model by adjusting the input, $\tilde{x}$, that is passed to the model 
so that, instead, it maximizes the loss function:
\begin{equation}
\max_{\tilde{x}\in x + \Delta} J(\theta,\tilde{x},y)
\label{eq:goal}
\end{equation}
Typically, these perturbations to $x$ cannot be arbitrary. In most scenarios, the generated example still needs to be close \enquote{enough} to the original input. In some contexts, certain features of $x$ cannot be edited or have to conform with the original input domain. We denote this allowable set of perturbations by $\Delta$. 

\subsection{Adversarial Attacks}
Typical classes of attacks in the literature are as follows. A white box attack assumes an attacker with full access to the model that can perform infinite queries, thus being able to devise strategies with high success rates \cite{carlini2017evaluating}. It has been argued, \cite{cheng2018queryefficient}, that such approaches are somewhat detached from reality: in practice, attackers have no access to the real model being attacked, and more importantly, there is a limited number of queries an attacker can perform before being flagged as suspicious and denied any further attempts. This latter approach has been coined as a \textit{\enquote{black-box}} attack, and has been studied in contexts that often resemble real world scenarios; i.e., when the attacker only has access to the model decision~\cite{cheng2018queryefficient}.  



Adversarial attacks aim to solve the maximization problem posed in Equation~\eqref{eq:goal}: given a set of allowed perturbations, find the adversarial sample that will maximize the loss of the target model. One of the first proposed methods to generate adversarial attacks was the Fast Gradient Sign Method (FGSM) \cite{goodfellow2015explaining}. FGSM solves the constrained problem by replacing the cost function $J$ with a linear approximation around the victim sample $x$. This solution can be interpreted as being a one-step scheme for solving Equation~\eqref{eq:goal} \cite{madry2017towards}. Later approaches proposed to  develop a stronger attack by iteratively perfecting the adversarial perturbation by performing projected gradient descent (PGD) upon the negative loss function. Considering FGSM as a single gradient step, PGD tries to solve Equation~\eqref{eq:goal} with multiple projected gradient steps, where the result is projected onto the set of valid adversarial examples \cite{madry2017towards}. 
For attacks on non-differentiable models or black-box settings, an approach is to compute zeroth order gradients to generate attacks~\cite{cheng2018queryefficient}.


\subsection{Adversarial Training}
In order to increase the robustness of classifiers against adversarial examples, Szegedy et al. \cite{szegedy2014intriguing} propose \textit{adversarial training}. 
It consists of a form of data augmentation where the goal is to expose flaws in how the classifier models its decision function by generating examples that target these specific sub-spaces of the input space \cite{goodfellow2015explaining}.
When training the model, adversarial examples are generated iteratively. The primary reason for this to be done at train time is that the concept of an adversarial example depends on the parameters of the model. An adversarial example generated prior to training is, most likely, very different to one generated post training. The authors show that by training the model using a mixture between clean and adversarial samples the model can thus be regularized~\cite{szegedy2014intriguing,goodfellow2015explaining}.

%% file: conclusions.tex
\section{Conclusions}\label{sec:conclusions}

In this work, we have presented a framework to train adversarially robust models on tabular data with focus on the fraud detection use case. We defined a perturbation space with a norm, as well as efficient methods to propagate perturbations to the feature engineered space of the model. Then we studied several attack search methods on such space, benchmarking them, and included the best method in the adversarial training loop to successfully developed models able to withstand strong attacks. This included studies of the best adversarial hyperparameters and, finally, an evaluation on test data to understand the trade-offs between model performance on clean data and attacked data for the adversarially trained model and the baseline. Our main conclusions are:
\begin{itemize}
    \item 
    Regarding the attack search method, we found greedy search to be the most effective.     
    \item Adversarial training prevents large performance drops (of about 30\%) against moderately large attacks and is essential against very aggressive  attacks (total loss in performance).
    \item Even for adversarially trained models with very strong attacks, their drop in performance on clean test data (with no attacks) is not larger than 7\%.
\end{itemize}
Interesting directions for future work are as follows. Our exploration of the hyperparameters was also mostly focused on parameters related to the adversarial components. Thus, it would be interesting to perform a more extensive search study of all the hyperparameters. Regarding comparisons with other attack search settings, we could also compare our results against a white box adversarial training setting and explore further attack search algorithms. As for the data, it would be interesting to study further datasets and use cases. It would also be interesting to be able to adversarially train a model
using a dataset that has surely undergone an adversarial change to test the robustness of a model adversarially trained before such a change happened.